\documentclass[runningheads]{llncs}
\usepackage{booktabs}
\usepackage{xcolor}
\usepackage{pifont}

\usepackage{listings}
\usepackage{comment}
\usepackage{xcolor} % Optional: For syntax highlighting colors

\usepackage{float}
\usepackage{enumitem}
\usepackage{amssymb}
\usepackage{algorithm}
\usepackage{algorithmic}
\usepackage{caption}
\usepackage[colorlinks=true,
            linkcolor=blue,
            urlcolor=blue,
            citecolor=blue]{hyperref}

\usepackage[T1]{fontenc}

\usepackage{graphicx}
\usepackage{booktabs}
\usepackage{amsmath}
\usepackage{pifont}
\usepackage{multirow}

\begin{document}
\title{ConformaDecompose: Explaining Uncertainty via Calibration Localization\thanks{This is an accepted author version of a paper to appear in the proceedings of the World Explainable AI Conference (Springer). The final authenticated version will be available via Springer.}}

\author{Fatima Rabia Yapicioglu\textsuperscript{1,2} \and Meltem Aksoy\textsuperscript{3} \and Alberto Rigenti\textsuperscript{2} \and Tuwe Löfström-Cavallin\textsuperscript{4}  \and Helena Löfström-Cavallin\textsuperscript{5} \and Şeyda Yoncacı\textsuperscript{6}  \and Luca Longo\textsuperscript{7}}

\authorrunning{F. R. Yapicioglu and M. Aksoy et al.}

 \institute{
Department of Computer Science and Engineering, University of Bologna, Italy
\and
Marketing and Sales, Automobili Lamborghini S.p.A, Sant'Agata Bolognese, Italy
\and
RCT Data Science and Security, Technical University Dortmund, Germany
\and
Jönköping AI Lab, Department of Computing, Jönköping University, Sweden
\and
Jönköping International Business School, Sweden
\and
Department of Software Engineering, Haliç University, Istanbul, Turkey
\and
University College Cork, Ireland
}

\maketitle              

\begin{abstract}

Conformal Prediction provides distribution-free prediction intervals with guaranteed coverage, but its reliance on a single global calibration threshold obscures the sources of uncertainty at the instance level. In particular, it conflates irreducible noise with uncertainty induced by heterogeneous training data (aleatoric), model limitations, or calibration mismatch (epistemic), offering little insight into why an interval is wide or whether it could be reduced. We introduce an \emph{uncertainty-aware explainability framework} that analyses the \emph{reducibility of calibration-induced epistemic conformal uncertainty} via progressive calibration localisation for regression tasks. The approach is diagnostic rather than causal: it does not estimate true aleatoric or epistemic uncertainty, but explains how conformal intervals contract and stabilise as calibration support is localised around a test instance. Across benchmarks and real-world data, absolute reducible uncertainty aligns with epistemic proxies, while its relative contribution varies by task, revealing regimes hidden by interval width. This instance-level view complements conformal uncertainty, enhancing interpretability without altering the predictor or coverage.

\end{abstract}

\section{Introduction}

Conformal Prediction (CP) provides a model-agnostic framework for predictive uncertainty quantification with distribution-free, finite-sample coverage guarantees under the assumption of exchangeability \cite{vovk2005algorithmic}. Given a calibration set, a held-out collection of examples used to assess model errors, and a user-specified miscoverage rate~$\alpha$, CP determines the prediction interval width by selecting the $(1-\alpha)$-quantile of calibration residuals. For example, in a sales-forecasting scenario, setting $\alpha = 0.05$ yields the 95th-percentile residual, which specifies the width of the interval around a point forecast. If a model predicts $120$ units for the next day and CP returns a $95\%$ interval of $[110,135]$, the resulting range provides a transparent and actionable measure of uncertainty for inventory and supply-chain planning. All coverage guarantees discussed in CP refer to marginal coverage under the standard exchangeability assumption \cite{angelopoulos2023conformal}.

Beyond coverage guarantees, uncertainty quantification research increasingly emphasises the distinction between aleatoric and epistemic uncertainty \cite{kendall2017uncertainties,lakshminarayanan2017simple}. In contrast, epistemic uncertainty arises from limited knowledge or model misspecification and is reducible, while aleatoric uncertainty reflects irreducible data noise. This distinction can be illustrated through a sales forecasting scenario. Demand prediction for a long-established product may exhibit uncertainty primarily driven by routine market fluctuations, reflecting aleatoric effects. In contrast, forecasts for a newly introduced product may be substantially more uncertain, not because demand is inherently noisier, but because little prior experience is available to inform the model, leading to uncertainty stemming from limited knowledge rather than natural variability. Importantly, this distinction is context-dependent: uncertainty that appears irreducible under a given model or representation may become reducible as more informative data or improved modelling assumptions become available \cite{hullermeier2021aleatoric}.

However, the global calibration mechanism of CP necessarily aggregates heterogeneous residual behaviour across the input space into a single calibration threshold \cite{javanmardi2024conformalized}. 
While this aggregation preserves valid marginal coverage, it conflates intrinsic aleatoric variability with epistemic effects arising from model limitations, sparse local support, or calibration mismatch. As a result, wide prediction intervals provide no instance-level insight into \emph{why} uncertainty is large, nor whether it could be reduced through more localised calibration. Moreover, even under ideal localisation, some epistemic uncertainty may persist due to model misspecification or limited expressiveness. These limitations motivate uncertainty-aware explainability approaches that go beyond interval width and explicitly reveal how predictive uncertainty forms, contracts, and stabilizes across heterogeneous calibration structures.

Prior work on CP has primarily focused on improving adaptivity to aleatoric variability and heteroscedastic noise by modifying nonconformity scores or enriching predictive models. Representative approaches include normalized CP~\cite{papadopoulos2008normalized}, conformalized quantile regression and its extensions~\cite{romano2019conformalized,rossellini2024integrating}. While these methods yield instance-dependent interval widths, uncertainty remains a single-shot outcome of calibration: they do not reveal how different regions of the calibration data contribute to the resulting interval width, nor do they indicate which portions of uncertainty could be reduced through localization. As a result, instance-wise interpretability of prediction intervals remains limited.

To address this gap, we present an uncertainty-aware explainability mechanism built on gradual \emph{calibration localization} for regression tasks. Instead of discarding calibration points or starting from a strictly local subset, we begin with the standard CP interval calibrated on the \emph{entire} calibration set, and progressively localize the calibration support through a distance-informed reweighting scheme. Calibration samples are clustered and softly downweighted in a structured global-to-local manner. 
%using a model-aware representation that defines similarity in terms of local model behavior, jointly capturing input geometry, prediction level, and model uncertainty. 
A greedy accept–revert rule ensures that the interval width decreases monotonically along the accepted localisation path, yielding a sequence of conformal-style localised intervals used solely for explanatory analysis on each test instance.

\begin{figure}[htbp]
    \centering
    \includegraphics[width=1\linewidth]{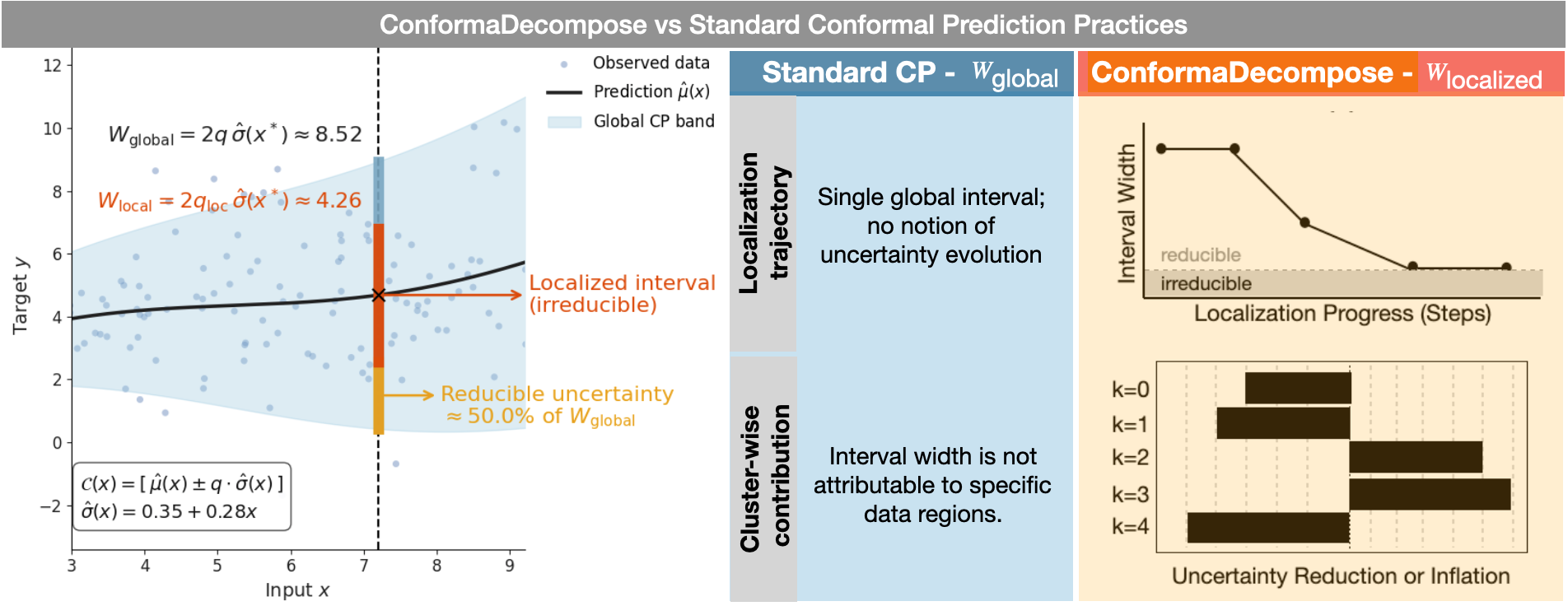}
    \caption{\textbf{Uncertainty decomposition on a toy regression task.} Synthetic data are generated as $y=\mu(x)+\varepsilon$, with $\mu(x)=2.0+0.45x+0.6\sin(0.6x)$ and $\varepsilon\sim\mathcal{N}(0,\sigma(x)^2)$, where $\sigma(x)=\textit{base}+\textit{slope}\cdot x$ (\textit{base}$=0.35$, \textit{slope}$=0.28$). The shaded band depicts a global conformal-style interval $\mathcal{C}(x)=[\hat{\mu}(x)\pm q_{\mathrm{global}}\hat{\sigma}(x)]$. At a query point $x^*$, the global interval width is decomposed into a reducible component and a remaining localised interval ($q_{\mathrm{loc}}<q_{\mathrm{global}}$), with the reducible fraction reported for explanation. The right panels contrast standard conformal prediction with \emph{ConformaDecompose}, illustrating how localisation attributes uncertainty reduction to specific calibration regions. No coverage guarantees are claimed.
}

    \label{fig:trajectory}
\end{figure}

Figure~\ref{fig:trajectory} contrasts standard conformal prediction, which yields a single global uncertainty interval, with \emph{ConformaDecompose}, which exposes uncertainty as a localisation process, revealing how interval width contracts and stabilises as irrelevant calibration regions are suppressed, enabling instance-wise attribution of global versus local uncertainty effects.

Taken together, this perspective reframes conformal uncertainty from a single scalar outcome into an interpretable process, making explicit the calibration-induced epistemic uncertainty that is reducible and revealing how heterogeneous regions of the calibration set contribute to the resulting prediction interval.

%The resulting \emph{uncertainty localization trajectory} shows how prediction intervals evolve as heterogeneous calibration regions are progressively downweighted. As illustrated in Figure~\ref{fig:trajectory}, panel~(A) depicts whether and where interval width contracts along the localization path, panel~(B) quantifies the reduction from the global to the most localized interval, panel~(C) indicates which calibration regions remain influential during localization, and panel~(D) attributes interval contraction or inflation to individual calibration clusters.

%Together, these views provide an instance-wise decomposition of predictive uncertainty into heterogeneity-induced global inflation, a reducible component driven by non-exchangeable calibration structure, and a stable regime of locally irreducible aleatoric noise, making explicit how different calibration regions shape uncertainty beyond a single scalar interval width.

Consider a scenario in an automotive residual value estimation setting with strong market and trim heterogeneity, our framework shows how global uncertainty is inflated by cross-segment mixing, how conditioning on relevant calibration support contracts intervals, and where irreducible noise remains. This decomposition informs pricing teams when uncertainty is noise-dominated, when additional data can reduce it, and when heterogeneity undermines predictive reliability. 

Our contributions are summarised as follows:
\begin{itemize}
\item \textit{Instance-wise uncertainty explainability via calibration localisation :}
We introduce \emph{ConformaDecompose}, a model-agnostic framework that localises and reweights calibration data based on model-aware representations, traces the contraction of conformal-style intervals, and enables an empirical decomposition of calibration-induced epistemic uncertainty into reducible and persistent components.

\item \textit{Multi-view diagnostics and empirical validation:}
We provide diagnostics for uncertainty localization, calibration, and cluster-wise interval contraction, showing consistent performance across datasets and granularities versus conformal baselines.

\item \textit{Practical relevance in real-world settings:}
We apply the framework to an automotive residual value estimation task, illustrating its interpretability and applicability under heterogeneous, real-world calibration conditions.

\end{itemize}

The remainder of this research article is organised as follows. Sections~2 and~3 review uncertainty types and conformal prediction methods. Section~4 introduces the proposed framework. Sections~5–7 present demonstrations, experimental results, and discussion. Section~8 concludes the paper.

\section{Preliminaries and Related Work}

\subsection{Uncertainty Quantification and Conformal Prediction}
Predictive uncertainty quantifies the reliability of a model’s predictions and supports informed decision-making by end users. It reflects an absolute measure of confidence and may arise from different sources throughout the prediction process, commonly distinguished as reducible uncertainty (often associated with epistemic uncertainty) and irreducible uncertainty (often associated with aleatoric uncertainty). Table \ref{tab:uncertainty_sources} summarises sources of predictive uncertainty across pre-modelling (data and sampling), modelling, and post-modelling stages; in this work, we focus on the epistemic uncertainty introduced during post-training calibration \cite{lofstrom2026concerning}.

\newcommand{\cmark}{\ding{51}} % check
\newcommand{\xmark}{\ding{55}} % cross

\begin{table}[htbp]
\centering
\caption{\textbf{Sources of uncertainty across the machine learning pipeline.}}
\label{tab:uncertainty_sources}
\begin{tabular}{p{3.8cm} p{4.5cm} c c}
\toprule
\textbf{Pipeline Component} & \textbf{Source of Uncertainty} & \textbf{Reducible} & \textbf{Irreducible} \\
\midrule

\multicolumn{4}{l}{\textit{Pre-modelling phase}} \\
\midrule

Data-generating process 
& Inherent stochasticity and measurement noise 
& \xmark & \cmark \\

Sampling and coverage of the data 
& Limited, biased, or incomplete data coverage 
& \cmark & \xmark \\

Feature representation and labelling 
& Encoding choices, annotation errors, or label ambiguity 
& \cmark & \cmark \\

\midrule
\multicolumn{4}{l}{\textit{Modelling phase}} \\
\midrule

Model specification and training 
& Model mismatch, optimisation instability, or capacity limitations 
& \cmark & \xmark \\

\midrule
\multicolumn{4}{l}{\textit{Post-modelling phase}} \\
\midrule

Distribution shift at deployment 
& Domain or population shift over time 
& \cmark & \xmark \\

Calibration-induced and uncertainty mapping 
& Imperfect calibration or uncertainty quantification mapping 
& \cmark & \xmark \\

\bottomrule
\end{tabular}
\end{table}

Conformal Prediction is a distribution-free framework for uncertainty quantification that constructs prediction sets with finite-sample marginal coverage guarantees under the assumption
of data exchangeability \cite{angelopoulos2023conformal}. Given a dataset
$\mathcal{D} = \{(x_i, y_i)\}_{i=1}^n$, the data are split into a training set
$\mathcal{D}_{\mathrm{train}}$ and a calibration set $\mathcal{D}_{\mathrm{cal}}$.
A predictive model $\hat f$ is trained on $\mathcal{D}_{\mathrm{train}}$, and
nonconformity scores are computed on the calibration set as
\[
s_i = A(y_i, \hat f(x_i)), \quad (x_i, y_i) \in \mathcal{D}_{\mathrm{cal}},
\]
where $A(\cdot)$ is a chosen nonconformity function.
For a target miscoverage level $\alpha \in (0,1)$, the $(1-\alpha)$ empirical quantile
$q_{1-\alpha}$ of the calibration scores is used to construct the prediction set for a new
input $x^\ast$ as
\[
\mathcal{C}(x^\ast) = \{y : A(y, \hat f(x^\ast)) \le q_{1-\alpha}\},
\]
which guarantees marginal coverage of at least $1-\alpha$ without assumptions on the data
distribution.

While this global calibration strategy ensures valid coverage, it aggregates heterogeneous residual behaviour across the input space into a single threshold, yielding prediction
intervals that are statistically valid but opaque. In particular, it obscures how uncertainty
at a given instance is influenced by calibration samples that are behaviorally irrelevant or
non-exchangeable with respect to that instance. To characterise the portion of predictive
uncertainty attributable to such calibration-induced aggregation effects, we introduce in Definition \ref{def:calib_epistemic}.

\begin{definition}[Calibration-induced epistemic conformal uncertainty]
\label{def:calib_epistemic}
Let $C_0(x^*)$ denote the conformal prediction interval for a test instance $x^*$ obtained under \emph{global calibration}, and let
\[
W_0(x^*) := \mathrm{width}\big(C_0(x^*)\big)
\]
be its interval width. Consider a sequence of localised intervals $\{C_t(x^*)\}_{t=0}^T$ obtained by progressively localising or reweighting the calibration set, and define
\[
W_{\min}(x^*) := \min_t \mathrm{width}\big(C_t(x^*)\big).
\]
The \emph{calibration-induced epistemic uncertainty} at $x^*$ is defined as
\[
W_{\mathrm{red}}(x^*) := W_0(x^*) - W_{\min}(x^*),
\]
capturing the portion of predictive uncertainty attributable to heterogeneous or behaviorally irrelevant calibration support under a fixed predictive model. The remaining width $W_{\min}(x^*)$ corresponds to uncertainty that cannot be further reduced through calibration localisation.
\end{definition}

\subsection{Uncertainty-Aware Explainability}

Uncertainty-aware explainability has recently emerged as a bridge between model interpretability and uncertainty quantification, and prior work can be grouped into two main research directions:
\begin{enumerate}[label=(\roman*)]
    \item \emph{Using uncertainty to explain model behaviour.}  
    In this direction, predictive uncertainty is used as an explanatory signal indicating model confidence and fragility \cite{yapicioglu2025conformasegment}. Representative approaches frame uncertainty estimation as XAI for local explanations and human-in-the-loop decisions under distribution shifts \cite{thuy2024explainability}, or propagate calibrated uncertainty through feature importance in regression and probabilistic regression \cite{lofstrom2025calibrated,bostrom2021mondrian}.

    \item \emph{Assessing the uncertainty of explanations themselves.}  
    This line of work evaluates explanation robustness by quantifying sensitivity to perturbations and by modelling explanation variance analytically or via attribution distributions \cite{chiaburu2025uncertainty,mulye2025uncertainty}, focusing on reliability rather than treating uncertainty as an explanatory signal.
\end{enumerate}
In this work, we focus on the first direction; we show how variations in interval width expose limitations of the calibration set, particularly distributional mismatch. In this sense, uncertainty serves as a diagnostic tool for attributing unreliability to calibration-induced sources rather than to model confidence alone.

\subsection{Related Work}

Several uncertainty-aware extensions of CP introduce adaptivity by modifying the nonconformity score or enriching the predictive model to encode local noise or uncertainty structure. 
Normalised conformal prediction \cite{papadopoulos2008normalized} rescales residuals using estimates of conditional heteroscedasticity, while conformalized quantile regression (CQR) \cite{romano2019conformalized} and uncertainty-aware CQR (UACQR) \cite{rossellini2024integrating} leverage conditional quantiles or ensemble dispersion to capture structured variability in the predictive distribution. 
Mondrian conformal prediction \cite{vovk2005algorithmic} introduces locality by partitioning the calibration set into predefined regimes and enforcing coverage within each partition, thereby enabling distribution-sensitive calibration without altering the underlying nonconformity score. 
%More recently, EPICSCORE \cite{cabezas2025epistemic} replaces the global calibration threshold with instance-dependent score distributions, increasing sensitivity to epistemic effects. 
A comparative summary of these approaches, together with the residual regimes in which they are applicable, is provided in Table~\ref{tab:comparison}.

\begin{table}[htbp]
\caption{\textbf{Comparison of uncertainty-aware conformal prediction approaches.}
Methods are compared by the uncertainty type addressed, instance-wise adaptivity,
explainability, and the residual regimes in which they are appropriate:
(i) homogeneous, (ii) heteroscedastic noise with fixed model error,
(iii) varying noise and model error \cite{sale2025aleatoric}.}

\vspace{1mm}
\centering
\small
\setlength{\tabcolsep}{4pt}
\renewcommand{\arraystretch}{1.2}
\begin{tabular}{p{2.2cm} p{2cm} c p{2.2cm} c c}
\hline
\textbf{Approach} 
& \textbf{Uncertainty} 
& \textbf{Ins.-wise} 
& \textbf{Mechanism} 
& \textbf{XAI} 
& \textbf{Reg.} \\
\hline

Normalized CP \cite{papadopoulos2008normalized} 
& Mixed 
& -- 
& Global quantile 
& -- 
& (i) \\
\hline

Mondrian \cite{vovk2005algorithmic}
& Mixed 
& \checkmark 
& Stratified quantile 
& Implicit 
& (i)--(ii) \\
\hline

CQR \cite{romano2019conformalized}
& Aleatoric
& \checkmark 
& Conditional quantile 
& Implicit 
& (ii) \\
\hline

CQR-r \cite{sesia2020comparison}
& Aleatoric 
& \checkmark 
& Width scaling 
& Implicit 
& (ii) \\
\hline

%EPICSC. \cite{cabezas2025epistemic}
%& Epistemic 
%& \checkmark 
%& Local score distribution 
%& Partial 
%& (iii) \\
%\hline

UACQR-P \\ UACQR-S \cite{rossellini2024integrating}
& Epistemic 
& \checkmark 
& UC-aware quantile 
& Implicit 
& (iii) \\
\hline

\textbf{ConformaDec.} 
& \textbf{Calibration-induced (Epistemic)} 
& \textbf{\checkmark} 
& \textbf{Calibration Localization} 
& \textbf{Explicit} 
& \textbf{(iii)} \\
\hline
\end{tabular}
\label{tab:comparison}
\end{table}

Across conformal approaches, uncertainty behavior can be summarized by three residual regimes \cite{sale2025aleatoric}: (i) homogeneous settings with exchangeable residuals and global thresholds; (ii) heteroscedastic noise with fixed model error, where local variance dominates and normalization-based methods primarily address aleatoric effects; and (iii) realistic settings with jointly varying noise and model error, yielding non-exchangeable residuals that confound heterogeneous uncertainty sources under a single global calibration rule. While methods such as CQR and UACQR produce instance-dependent prediction intervals, uncertainty remains implicitly encoded and collapsed into a single scalar width, obscuring how calibration support contributes to uncertainty or whether further localisation could reduce it. By contrast, \emph{ConformaDecompose} performs post-hoc calibration localisation, explicitly exposing how uncertainty evolves as calibration support is progressively restricted.

%Despite these advances, all existing variants ultimately operate by modifying the nonconformity score or the predictive model, and consequently treat predictive uncertainty as a single-shot outcome summarized by a scalar interval width. The calibration process itself remains largely opaque: practitioners are not informed how residuals from different regions of the calibration set contribute to uncertainty, nor whether a wide interval reflects intrinsic noise, limited local support, or model unfamiliarity. As a result, instance-wise uncertainty decomposition remains ill-posed, and users lack an explanation of how uncertainty accumulates and stabilizes across heterogeneous calibration structure. This limitation motivates moving beyond adaptive interval construction toward an uncertainty-aware explainability framework that explicitly reveals how uncertainty forms, contracts, and stabilizes as calibration support is progressively localized.

\section{Problem Formulation and Presenting Our Approach}

\subsection{Limitations of Global Calibration and Ill-Posedness of Instance-Wise Uncertainty}

We consider a regression setting in which observations follow $Y=f(X)+\varepsilon$, where $\varepsilon$ captures aleatoric noise and discrepancies between $f$ and its estimator $\hat f$ reflect epistemic effects. Given a trained predictor $\hat f$, CP constructs prediction intervals by aggregating calibration residuals $R=|Y-\hat f(X)|$ into a single exchangeable set and selecting a global $(1-\alpha)$ quantile $q_{1-\alpha}$, yielding intervals of the form $\hat f(x)\pm q_{1-\alpha}$. While this procedure guarantees distribution-free coverage, global calibration implicitly mixes heterogeneous residual behaviour across the input space. The resulting residual distribution can be viewed as a mixture of local residual distributions, conflating intrinsic noise, model error, and region-specific calibration effects. Consequently, the global threshold may be dominated by calibration regions irrelevant to a given test instance, thereby obscuring the local uncertainty structure.

This aggregation renders instance-wise uncertainty decomposition ill-posed under standard CP: interval width alone cannot distinguish between locally irreducible uncertainty and uncertainty that could be reduced through localised calibration. Although adaptive variants improve local calibration by addressing heteroscedastic noise, epistemic effects remain globally aggregated, leaving uncertainty purely descriptive.

Consequently, existing conformal frameworks indicate the width of an interval, but not why it is wide or whether further localisation could reduce it, motivating uncertainty-aware explainability mechanisms that expose how uncertainty emerges, contracts, and stabilises under heterogeneous calibration support.

\subsection{Open Questions and Motivation}

Instead, uncertainty is produced as a single scalar outcome, offering limited insight into the mechanisms through which different regions of the calibration set influence the final interval.

Therefore, predictive uncertainty across existing conformal approaches is computed in a one-off, obscuring two key questions:
\begin{enumerate}[label=(\roman*), leftmargin=1.5em]
    \item which portions of predictive uncertainty arise from heterogeneous calibration structure (calibration-induced epistemic conformal uncertainty, Definition \ref{def:calib_epistemic}) and limited local support (due to model/data limitations), and
    \item at what point uncertainty stabilises and becomes irreducible, reflecting locally unavoidable noise.
\end{enumerate}
Addressing these questions requires moving beyond merely adapting interval width toward an explainability framework that explicitly reveals how uncertainty forms, contracts, and stabilises as calibration support is progressively localised. This gap motivates the introduction of the uncertainty localisation introduced in this work.

\subsection{ConformaDecompose Structure and Working Mechanism}

In heterogeneous regression settings, a single globally calibrated conformal interval is sufficient for coverage but limited for explanation. \emph{ConformaDecompose} addresses this limitation by transforming conformal uncertainty from a single-shot output into an instance-wise explainability process. The framework is structured around the following guiding technical enabler questions for a test instance $x^*$:
\begin{enumerate}[label=\textbf{Q\arabic*:}, leftmargin=2.2em]
    \item What predictive reference (prediction and uncertainty proxy) anchors the analysis for $x^*$?
    \item Which calibration samples are behaviorally relevant to $x^*$, and how does uncertainty evolve as less relevant regions are progressively downweighted?
    \item When does interval contraction stabilise, and how much of the global uncertainty is structurally reducible versus persistent under localisation?
\end{enumerate}
These questions motivate the components described next, which together yield an interpretable localisation trajectory and an empirical decomposition of conformal uncertainty.

\noindent
\subsubsection{Establishing a local model instability proxy and nonconformity score for localisation.}

Before constructing the model-aware calibration space, we define a fixed, instance-specific \emph{local model instability proxy} (referring to Q1) as formalised in Definition~\ref{sigma}, together with the nonconformity score in Equation~(\ref{nc}). This proxy serves as a common reference for both the construction of the calibration space and the subsequent diagnostic analysis.

The scale $\sigma^*$ captures model-dependent instability arising from
limited local training support, model sensitivity, or structural
misspecification of the predictor. It reflects how fragile the model’s
prediction is at a given test instance, rather than the uncertainty induced by the calibration procedure itself. Importantly, $\sigma^*$ is computed once per test instance and remains fixed throughout the calibration localisation process, ensuring that variations in the prediction interval width are driven solely by changes in the calibration support rather than by model adaptation.

We emphasize that $\sigma^*$ is \emph{not} intended to estimate
calibration-induced epistemic uncertainty. Instead, it provides an independent signal of local model fragility against which the behaviour of uncertainty reducibility can be examined. Unlike aleatoric noise, $\sigma^*$ varies across instances but remains invariant under localisation, allowing us to assess whether and how calibration-induced uncertainty responds to regions where the model itself is epistemically fragile. All relationships examined between $\sigma^*$ and reducible uncertainty are therefore associative rather than causal.

\begin{definition}[Local model instability proxy ($\sigma^*$)]
\label{sigma}
Let $x^*$ denote a test instance and let $\mu^* = \hat f(x^*)$ be the
corresponding point prediction. Let $\hat\sigma(\cdot)$ denote a
model-specific uncertainty or instability estimation function. We define the
local model instability proxy for $x^*$ as
\begin{equation}
\sigma^* \;\triangleq\; \hat\sigma(x^*).
\end{equation}
For each test instance $x^*$, the pair $(\mu^*, \sigma^*)$ defines a fixed predictive reference. 
\end{definition}

In practice, $\hat\sigma(\cdot)$ is instantiated using model-specific routines.
For Random Forests, $\hat\sigma(x^*)$ is computed as the standard deviation of predictions across individual trees, providing an estimate of ensemble prediction variability. This quantity is computed prior to calibration and treated as fixed throughout the localisation procedure.

For Ridge Regression, an auxiliary regressor $\hat g(\cdot)$ is fitted to the absolute calibration residuals
\[
r_i = |y_i - \hat f(x_i)|,
\]
to approximate the expected residual magnitude conditional on $x$. The resulting estimate is evaluated at $x^*$ and stabilised via
\[
\hat\sigma(x^*) \leftarrow \max\!\big(\hat g(x^*),\, \varepsilon\big),
\]
where $\varepsilon > 0$ prevents degenerate or numerically unstable variance estimates.

Conformal prediction does not prescribe a unique uncertainty scalar; any instance-dependent scale may be used, provided it is held fixed during calibration. We leverage this flexibility to define a model-aware, instance-level epistemic baseline, ensuring that the normalised nonconformity score
\begin{equation}
\label{nc}
s^* = \frac{\lvert y^*_{\mathrm{true}} - \mu^* \rvert}{\sigma^*}
\end{equation}
remains invariant throughout localisation. At inference time, the proposed localisation and interval construction rely only on $(\mu^*, \sigma^*)$ and the calibration set, and do not require access to $y^*_{\mathrm{true}}$. Consequently, changes in the width of the prediction interval reflect variations in calibration support rather than model-induced variability.

\noindent
\subsubsection{Representing calibration data in a model-aware space.}

Let $(X_{\mathrm{cal}}, Y_{\mathrm{cal}})$ denote the calibration set, with
$X_{\mathrm{cal}} = \{x_i\}_{i=1}^{N_{\mathrm{cal}}}$ and
$Y_{\mathrm{cal}} = \{y_i\}_{i=1}^{N_{\mathrm{cal}}}$.
We construct normalised nonconformity scores
\begin{equation}
A(X_{\mathrm{cal}}, Y_{\mathrm{cal}})
= \left\{ \frac{|y_i - f(x_i)|}{\hat\sigma(x_i)} \right\}_{i=1}^{N_{\mathrm{cal}}},
\end{equation}
thereby aggregating residuals over the entire calibration set.
This aggregation effectively homogenises heterogeneous residual behaviour and obscures local structure in predictive uncertainty. To explicitly capture heterogeneity in model behaviour, we embed each calibration instance into a \emph{model-aware representation} that jointly encodes input geometry, model output, and predictive uncertainty.
Each calibration point $x_i$ is mapped to
\begin{equation}
z_i = \big[ \lambda_X x_i,\; \lambda_\mu f(x_i),\; \lambda_\sigma \hat\sigma(x_i) \big],
\label{zi}
\end{equation}
where $\hat\sigma(x_i)$ denotes the predictive scale at $x_i$, and
$\lambda_X$, $\lambda_\mu$, and $\lambda_\sigma$ control the relative influence of geometry, prediction level, and uncertainty, respectively.
This representation defines similarity in terms of local model behavior rather than raw feature proximity alone.

\noindent
Each calibration point is thus characterised by three complementary components, and two points are considered behaviorally similar \emph{only if} they align across all three:
\begin{itemize}
    \item \textbf{Input location} ($x_i$), ensuring that clustering respects the geometric structure of the input space;
    \item \textbf{Model prediction} ($f(x_i)$), capturing similarity in local predictive behaviour beyond mere proximity;
    \item \textbf{Predictive uncertainty} ($\hat\sigma(x_i)$), distinguishing well-supported regions from epistemically uncertain ones.
\end{itemize}

Algorithm~\ref{alg:clustering_order_compact} clusters calibration samples in this model-aware space and orders clusters by relevance to a test instance, enabling progressive localisation in which behaviorally distant regions are downweighted first. This yields monotonic interval contraction, separating uncertainty driven by global heterogeneity from locally persistent uncertainty and enabling instance-wise explanation. Clustering acts as a generic neighbourhood construction mechanism, performed once on the calibration set, allowing alternative similarity measures without altering the framework.

\begin{algorithm}[htbp]
\caption{Model-aware calibration clustering and ordering}
\label{alg:clustering_order_compact}
\begin{algorithmic}[1]
\REQUIRE 
$X_{\mathrm{cal}}=\{x_i\}_{i=1}^{N_{\mathrm{cal}}}$, test instance $x^*$, model $f$, scale proxy $\hat\sigma(\cdot)$, clusters $K$, weights $(\lambda_X,\lambda_\mu,\lambda_\sigma)$
\ENSURE 
Assignments $\mathbf{c}$, centroids $\{\bar z_k\}$, ordering $\pi$

\STATE $z_i \gets [\lambda_X x_i,\;\lambda_\mu f(x_i),\;\lambda_\sigma \hat\sigma(x_i)] \;\forall i$ \COMMENT{Embed calibration instances}

\STATE $z^* \gets [\lambda_X x^*,\;\lambda_\mu f(x^*),\;\lambda_\sigma \hat\sigma(x^*)]$ \COMMENT{Embed test instance}

\STATE $(\mathbf{c},\{\bar z_k\}) \gets \mathrm{KMeans}(\{z_i\};K)$ \COMMENT{Cluster behavioral similarity \cite{mcqueen1967some}}

\STATE $D_k \gets \|\bar z_k - z^*\|_2 \;\forall k$ \COMMENT{Cluster-test instance distance}

\STATE $\pi \gets \mathrm{argsort}(D_1,\ldots,D_K)$ \COMMENT{Sort clusters in distance decreasing order}

\RETURN $\mathbf{c}, \{\bar z_k\}, \pi$
\end{algorithmic}
\end{algorithm}

\noindent
\subsubsection{Localising calibration support without discarding data.}

To enable localisation, calibration clusters are ranked by behavioural relevance to the test embedding $z^*$, inducing a global-to-local hierarchy without hard sample selection (Algorithm~\ref{alg:clustering_order_compact}).

At each localisation step, uncertainty is computed via a soft reweighting mechanism rather than strict subset selection.
Reweighting is used as an explanatory probe of calibration support rather than to introduce new guarantees, and localised intervals are interpreted diagnostically to explain uncertainty.

Concretely, as described in Algorithm~\ref{alg:weighted_cp}, calibration samples from clusters currently deemed relevant receive unit weight, while samples from discarded clusters are downweighted by a small constant $\varepsilon > 0$.
%This ensures that all calibration points retain non-zero influence, thereby preserving the theoretical coverage guarantees of conformal prediction.
A relevance-aware conformal quantile is then computed using a weighted $(1-\alpha)$ quantile estimator, and prediction intervals are constructed around the fixed test forecast $(\mu^*, \sigma^*)$.

By progressively downweighting behaviorally distant calibration regions while maintaining a valid calibration distribution, the procedure yields a sequence of increasingly localised prediction intervals.
The resulting contraction in interval width quantifies how much predictive uncertainty arises from globally heterogeneous calibration behaviour versus locally irreducible uncertainty around the test instance.

\begin{algorithm}[htbp]
\caption{Weighted Conformal Interval Computation}
\label{alg:weighted_cp}
\begin{algorithmic}[1]
\REQUIRE
Calibration scores $\{r_i\}_{i=1}^{N_{\mathrm{cal}}}$,  
cluster assignments $\{c_i\}$,  
active cluster set $\mathcal{U}$,  
minimum weight $\varepsilon$,  
miscoverage level $\alpha$,  
test prediction $(\mu^*, \sigma^*)$
\ENSURE
Prediction interval $C(x^*)$

\vspace{1mm}

\FOR{$i = 1$ to $N_{\mathrm{cal}}$}
    \IF{$c_i \in \mathcal{U}$}
        \STATE $w_i \leftarrow 1$
        \COMMENT{Fully weight calibration points in active (relevant) clusters}
    \ELSE
        \STATE $w_i \leftarrow \varepsilon$
        \COMMENT{Downweight calibration points from discarded clusters}
    \ENDIF
\ENDFOR

\vspace{1mm}
\STATE Compute weighted $(1-\alpha)$-quantile
\[
\hat q_w \leftarrow 
\inf \left\{ q :
\frac{\sum_i w_i \mathbb{I}(r_i \le q)}{\sum_i w_i}
\ge 1-\alpha
\right\}
\]

\vspace{1mm}
%\COMMENT{Step 3: Construct prediction interval around anchored test forecast}
\STATE Construct conformal interval
\[
C(x^*) =
[\mu^* - \sigma^* \hat q_w,\;
 \mu^* + \sigma^* \hat q_w]
\] \COMMENT{Interval width adapts to retained calibration regions}

\RETURN $C(x^*)$
\end{algorithmic}
\end{algorithm}

\noindent
\subsubsection{Constructing an interpretable localisation trajectory.}
Given a weight vector $w$ over calibration samples, a weighted conformal quantile $\hat q_w$ is computed from the calibration nonconformity scores, yielding the corresponding prediction interval
\begin{equation}
C_w(x^*) = \left[ \mu^* - \sigma^* \hat q_w,\; \mu^* + \sigma^* \hat q_w \right],
\end{equation}
as detailed in Algorithm~\ref{alg:weighted_cp}.
This formulation allows the calibration distribution to be progressively localised around the test instance.

Algorithm~\ref{alg:greedy_path_compact} performs greedy calibration localisation by iteratively downweighting clusters based on test-instance relevance, accepting steps only if interval width does not increase. The resulting non-increasing interval trajectory reveals how uncertainty contracts as irrelevant calibration regions are suppressed, separating persistent uncertainty from calibration-induced heterogeneity.

\begin{algorithm}[htbp]
\caption{Greedy calibration localisation trajectory}
\label{alg:greedy_path_compact}
\begin{algorithmic}[1]
\REQUIRE Calibration data $(X_{\mathrm{cal}},Y_{\mathrm{cal}})$, test instance $x^*$, error level $\alpha$
\ENSURE Localization path $\mathcal{T}$, reducible uncertainty $R_{\mathrm{red}}$, cluster effects $\Delta \mathbf{W}$

\STATE $(\mathbf{c},\{\bar z_k\},\pi) \gets$ Alg.~\ref{alg:clustering_order_compact}
\COMMENT{Learn calibration structure and relevance ordering}

\STATE $(C_0,W_0) \gets$ Alg.~\ref{alg:weighted_cp} with $s=0$
\COMMENT{Global conformal interval (no localisation)}

\STATE $\mathcal{T} \gets \{(0,W_0)\}$,\quad $W_{\mathrm{cur}} \gets W_0$,\quad $\Delta \mathbf{W} \gets [\,]$
\COMMENT{Initialize localization path and baseline uncertainty}

\FOR{$s=1$ to $K-1$}
    \STATE $(C_s,W_s) \gets$ Alg.~\ref{alg:weighted_cp} with step $s$
    \COMMENT{Compute interval after removing $s$ least relevant clusters}

    \STATE Append $(W_s-W_{\mathrm{cur}})$ to $\Delta \mathbf{W}$
    \COMMENT{Measure uncertainty change from current configuration}

    \IF{$W_s \le W_{\mathrm{cur}}$}
        \STATE $W_{\mathrm{cur}} \gets W_s$
        \COMMENT{Accept localisation if uncertainty does not increase}
    \ENDIF

    \STATE Append $(s,W_{\mathrm{cur}})$ to $\mathcal{T}$
    \COMMENT{Record uncertainty along localisation trajectory}
\ENDFOR

\STATE $W_{\min} \gets \min_{(\cdot,W)\in\mathcal{T}} W$
\COMMENT{Most localised uncertainty level}

\STATE $R_{\mathrm{red}} \gets (W_0-W_{\min})/W_0$
\COMMENT{Relative reducible calibration-induced uncertainty}

\RETURN $\mathcal{T}, R_{\mathrm{red}}, \Delta \mathbf{W}$
\end{algorithmic}
\end{algorithm}

%FORMULIZE OUTPUT PROPERLY

\noindent
\subsubsection{Empirical decomposition of predictive uncertainty.}
The empirical decomposition of predictive uncertainty is defined through \emph{calibration-induced reducible uncertainty ratio} (Definition~\ref{def:Rred}). 
It captures interval width attributable to heterogeneous calibration structure and behaviorally irrelevant calibration regions, while the remaining stable width reflects uncertainty that persists under calibration localisation. Width-based metrics quantify the magnitude of reducible uncertainty, whereas ratio-based metrics characterise its relative composition. Together, they distinguish uncertainty strength from uncertainty structure.
This decomposition is empirical and diagnostic: irreducible uncertainty does not imply purely aleatoric noise, but rather uncertainty that cannot be further reduced by calibration localisation within a fixed predictive model.

\begin{comment}
\begin{definition}[Reducible Calibration-Induced Epistemic Conformal Uncertainty Width]
\label{def:Wred}
Let $W_{\text{global}}$ denote the width of the conformal prediction interval obtained under global calibration (using all regimes in the calibration set), and let $W_{\min}$ denote the minimum interval width attained along the localization trajectory. The reducible uncertainty is defined as
\begin{equation}
W_{\text{red}} = W_{\text{global}} - W_{\min}.
\end{equation}
\end{definition}
\end{comment}

\begin{definition}[Reducible Calibration-Induced Epistemic Conformal Uncertainty Ratio]
\label{def:Rred}
Let $W_{\text{global}}$ denote the width of the conformal prediction interval under global calibration, and let $W_{\min}$ denote the minimum interval width obtained via localised calibration via ConformaDecompose. 

Define the reducible calibration-induced uncertainty width as
\begin{equation}
W_{\text{red}} = W_{\text{global}} - W_{\min}.
\end{equation}

The reducible calibration-induced epistemic uncertainty ratio is then defined as
\begin{equation}
R_{\text{red}} = \frac{W_{\text{red}}}{W_{\text{global}}}.
\end{equation}
\end{definition}

\begin{figure}[htbp]
    \centering
    \includegraphics[width=0.9\linewidth]{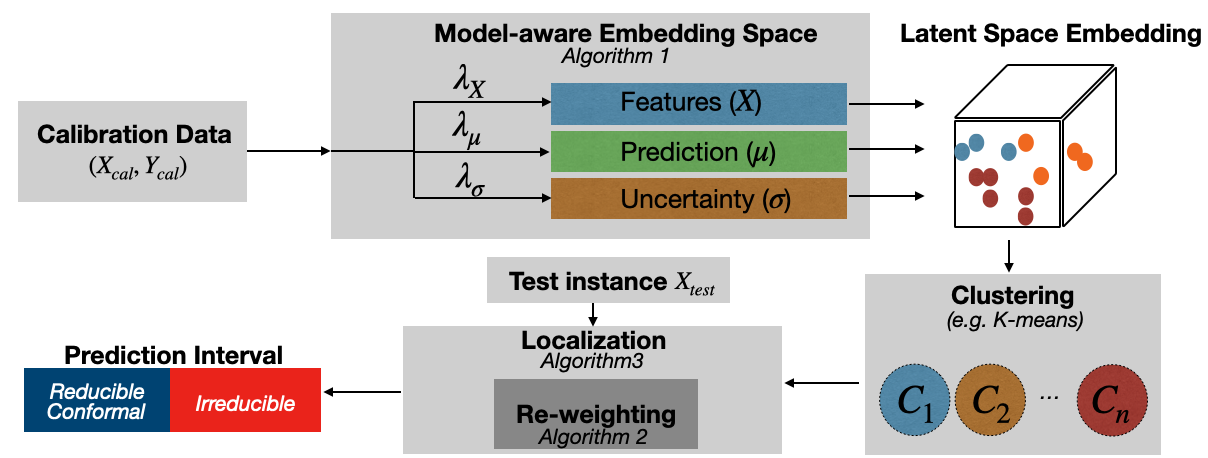}
\caption{\textbf{ConformaDecompose overview}. Calibration data $(X_{cal},Y_{cal})$ are embedded using weighted features $(\lambda_X)$, predictions $(\lambda_\mu)$, and uncertainty $(\lambda_\sigma)$, clustered $(C_1,\dots,C_n)$, and used to localize $X_{test}$.}
    \label{fig:sample-method}
\end{figure}

The overall decomposition (Figure~\ref{fig:sample-method}) is realised at the instance level through a greedy localisation trajectory (Algorithm~\ref{alg:greedy_path_compact}) and summarised by the four-panel diagnostic output in Figure~\ref{fig:sample-output}. The panels report: (i) the uncertainty localisation trajectory, showing how the prediction interval width evolves as calibration regions are progressively downweighted; (ii) the corresponding relative uncertainty reduction expressed as a fraction of the global interval width; (iii) cluster-wise contributions to uncertainty reducibility, indicating which calibration regions drive interval contraction; and (iv) a cluster usage heatmap, visualising how calibration regions are retained or suppressed across localisation steps.

\subsection{Efficiency and Computational Complexity}
Let $n = |{\cal D}_{\mathrm{cal}}|$ denote the calibration size, $d$ the input dimension, $K$ the number of clusters, and $p=d+2$ the dimensionality of the clustering space
$Z=[\lambda_X X,\lambda_\mu \mu,\lambda_\sigma \sigma]$.
Let $C_\mu(n)$ and $C_\sigma(n)$ denote the costs of computing $\mu(X_{\mathrm{cal}})$ and $\sigma(X_{\mathrm{cal}})$, and let $I$ and $n_{\mathrm{init}}$ be the number of Lloyd iterations and initializations in $k$-means.

The runtime is dominated by: (i) construction of $Z$, with cost $O(nd + C_\mu(n)+C_\sigma(n))$; (ii) $k$-means clustering, with cost $O(n_{\mathrm{init}} I n K p)$; and (iii) a greedy localization path of at most $T\le K-1$ steps, where each step computes a weighted conformal quantile in $O(n\log n)$ time. The overall time complexity is therefore
\[
O\!\left(C_\mu(n)+C_\sigma(n) + n_{\mathrm{init}} I n K p + T n \log n\right),
\qquad p=d+2,\; T\le K-1.
\]
The memory complexity is $O(np)$ for storing $Z_{\mathrm{cal}}$, plus $O(n)$ for calibration scores and weights, in addition to model storage.

\section{Illustrative Case Study: Craigslist Used Cars Dataset}
\subsection{Illustration of an instance-wise output} 

\begin{figure}[htbp]
    \centering
    \includegraphics[width=0.9\linewidth]{car_sale.png}
\caption{\textbf{Instance-level uncertainty-aware explainability via calibration localization ($K=4$).}
\textbf{(A)} Localisation path showing interval contraction from global to full localisation, reducing width from $4855.40$ to $4545.24$ \$.
\textbf{(B)} Absolute interval reduction of i.e., $6.39\%$ reducibility.
\textbf{(C)} Calibration support heatmap showing progressive downweighting of clusters and increasing alignment with the test instance.
\textbf{(D)} Cluster-wise width changes under cluster removal, interpreted relative to global calibration.}
    \label{fig:sample-output}
\end{figure}

To illustrate instance-level interpretability, we use a real-world used-car pricing dataset from online listings (the Craigslist Used Car Dataset \cite{reese2020craigslist}) solely for demonstration. For a selected test instance, \emph{ConformaDecompose} produces a four-panel diagnostic visual as shown in Figure~\ref{fig:sample-output}. 

We organised approximately $N \approx 73{,}245$ calibration listings and $N \approx 219{,}750$ training listings, with mixed numerical and categorical features, in which the listing price is modelled as a continuous regression target. After preprocessing (removal of incomplete records, categorical encoding, and normalisation), the data comprise approximately d = 10 effective features, such as vehicle age, mileage, manufacturer, model, fuel type, transmission, location, etc. A Random Forest regressor is adopted as the base model, and the number of clusters is set to $K=4$.

\noindent
\textbf{Test instance profile.}
The selected test instance corresponds to a relatively young, low-mileage, high-price vehicle, with a vehicle age of 7 years, mileage of 43,628 miles, and listing price of approximately \$40,300. This places the instance in a higher-value vehicle segment than the majority of calibration samples.

\noindent
\textbf{Calibration cluster structure and uncertainty contributions.}
Table~\ref{tab:demo_cluster_profiles_impact} summarises the characteristics of calibration clusters in terms of mean vehicle age, mileage, and price, and illustrates their impact on predictive uncertainty, thereby demonstrating how users can effectively interpret and utilise the model outputs.

\begin{table}[htbp]
\centering
\caption{\textbf{Calibration cluster profiles, market interpretation, and uncertainty impact.} Mean values are reported for each feature except ``Width Change''.}
\label{tab:demo_cluster_profiles_impact}
\begin{tabular}{c p{1.3cm} p{1.3cm} p{1.3cm} p{4.8cm} p{1.6cm}}
\toprule
\textbf{Cluster} & \textbf{Age} & \textbf{Mileage} & \textbf{Price} & \textbf{Interpretation} & \textbf{Width Change ($-\Delta$)} \\
\midrule
k=2 & 19.5 yrs & 217k & \$8k & Very old / high-mileage vehicles & -605.48 \\
k=0 & 16.3 yrs & 144k & \$11k & Older mid-range used vehicles & +168.72 \\
k=3 & 13.5 yrs & 86k & \$17.5k & Mid-age vehicle segment & -14.48 \\
k=1 & \textbf{9.5 yrs} & \textbf{26k} & \textbf{\$28.5k} & Newer, high-value segment & +141.44 \\
\bottomrule
\end{tabular}
\end{table}

Comparison with cluster statistics shows that Cluster~1 is most aligned with the test vehicle, exhibiting the closest age–mileage regime and the highest average price. In contrast, Clusters~2 and~0 represent older, higher-mileage, lower-price segments that are less compatible with the test instance.

During localisation, these incompatible clusters are progressively downweighted, while clusters aligned with the test instance remain active longer. \textit{ConformaDecompose} suppresses calibration samples that induce heterogeneous residual behaviour, isolating uncertainty arising from calibration cluster aggregation.

Cluster-wise width changes support this mechanism: negative values indicate uncertainty inflation under global calibration, while positive values reflect stabilising calibration support. Cluster~2 contributes the strongest negative effect, showing that very old, high-mileage vehicles substantially inflate uncertainty for this instance. Cluster~1 provides stabilising support consistent with the test vehicle, while Clusters~0 and~3 show intermediate compatibility.

Overall, the example illustrates that pricing uncertainty partly arises from mixing low-price, high-mileage vehicles with higher-end vehicles, and \textit{ConformaDecompose} enables identification of these heterogeneous calibration effects.

\subsection{There exists an optimal and interpretable calibration resolution}

As we summarised in Table \ref{tab:heterogeneity_reducibility}, calibration heterogeneity increases monotonically with clustering granularity, reflecting progressively finer segmentation of behavioural regimes in the calibration data (Spearman\cite{spearman1987proof}, $p < 0.05$).

\begin{table}[htbp]
\centering
\caption{\textbf{Calibration heterogeneity and calibration-induced reducible uncertainty across clustering granularities.} 
Heterogeneity is computed as $H = \mathrm{Var}(\{\sigma_k^2\}_{k=1}^{K})$, where $\sigma_k^2$ denotes the residual variance within cluster $k$.}
\label{tab:heterogeneity_reducibility}
\begin{tabular}{p{4cm} p{1cm} p{1.2cm} p{1.4cm} p{1.4cm} p{1.2cm}}
\toprule
\textbf{K} & 1 & 2 & 4 & 6 & 8 \\
\midrule
\textbf{Heterogeneity ($\times 10^6$)} & 0 & 5.74 & 7.66 & 7.92 & 8.63  \\
\textbf{Reducibility} & -- & -2.74\% & -6.39\% & \textbf{-23.21\%} & -18.76\% \\
\bottomrule
\end{tabular}
\end{table}

Reducible uncertainty initially increases with heterogeneity, indicating that coarse calibration partitions mix incompatible vehicle segments and inflate global conformal uncertainty. The largest reducibility occurs at moderate granularity ($K=6$), suggesting that this level best captures structurally meaningful calibration regimes. At higher granularity ($K=8$), reducibility decreases despite increasing heterogeneity, indicating that overly fine partitions fragment calibration support and reduce localisation efficiency.

From a practical perspective, these results demonstrate that the effectiveness of uncertainty localisation depends on calibration and structural resolution. Users can leverage this behaviour to diagnose whether predictive uncertainty arises primarily from heterogeneous calibration aggregation or from persistent data and model variability. In the context of used-vehicle pricing, the results indicate that uncertainty inflation partly arises from mixing low-price, high-mileage vehicles with higher-value, low-mileage vehicles. By identifying clustering resolutions that maximise reducible uncertainty, \textit{ConformaDecompose} provides a principled mechanism for selecting calibration structures that enhance interpretability and improve uncertainty diagnostics. For practitioners, \textit{ConformaDecompose} enables:

\begin{itemize}
    \item \textit{Calibration diagnostics}: detecting incompatible calibration regimes and guiding filtering, reweighting, or stratification.
    \item \textit{Resolution tuning}: selecting clustering granularity that balances localisation and calibration stability.
    \item \textit{Model--data mismatch detection}: separating aggregation-induced uncertainty from intrinsic model variability.
    \item \textit{Robustness monitoring}: auditing instability of prediction intervals under regime shifts or deployment drift.
\end{itemize}

Implementation and schematic are available at the following GitHub Repository: \url{https://github.com/rabia174/ConformaDecompose}.

\section{Experiments and Evaluations}
\subsection{Experimental Settings}
\subsubsection{\textbf{Datasets.}} The datasets in Table~\ref{tab:datasets} cover diverse domains and noise characteristics, enabling evaluation of uncertainty reducibility across heterogeneous settings. This diversity allows us to examine how the proposed localisation trajectory behaves under both aleatoric- and epistemic-dominated regimes, rather than being tailored to a specific data distribution.

\begin{table}[htbp]
\caption{\textbf{Datasets used for evaluation.}
Summary of datasets used in the experiments, including targets, dimensionality, sample size, and application domain.}
\label{tab:datasets}

\vspace{1mm}
\centering
\small
\setlength{\tabcolsep}{4pt}
\renewcommand{\arraystretch}{1.2}
\begin{tabular}{p{4.2cm} p{2.2cm} c c p{2.4cm}}
\toprule
\textbf{Dataset} &
\textbf{Target} &
\textbf{Feat.} &
\textbf{Samples} &
\textbf{Domain} \\
\midrule

%Wine Quality \cite{cortez2009winequality} &
%Quality score &
%11 &
%4{,}898 &
%Food quality \\

Combined Cycle Power P. \cite{tfekci2014ccpp}&
Energy output &
4 &
9{,}568 &
Energy systems \\

Parkinson's Telemonit. \cite{tsanas2009parkinsonstelemonitoring}&
Motor UPDRS &
16 &
5{,}875 &
Healthcare \\

%Seoul Bike Sharing %\cite{seoulbike2020sharingdemand}&
%Bike count &
%11 &
%8{,}760 &
%Urban mobility \\

Airfoil Self-Noise \cite{brooks1989airfoilselfnoise} &
Sound pressure &
5 &
1{,}503 &
Aeroacoustics \\

\begin{comment}
   Naval Propulsion CBM \cite{condition_based_maintenance_of_naval_propulsion_plants_316}&
Comp. decay &
16 &
11{,}934 &
Indust. maint. \\ 
\end{comment}

Abalone \cite{abalone_1}&
Rings (age) &
8 &
4{,}177 &
Biology \\

\bottomrule
\end{tabular}
\end{table}

\subsubsection{\textbf{Models.}} To evaluate the proposed uncertainty decomposition across models with distinct inductive biases, we employ Random Forest and Ridge Regression. Random Forests are non-linear, ensemble-based learners that capture complex interactions and heterogeneous data regimes, whereas Ridge Regression provides a linear, regularised baseline with stable and interpretable behaviour. This pairing enables assessment of uncertainty–reducibility relationships under both flexible and constrained modelling assumptions, allowing model-dependent effects to be examined without introducing architectural complexity as summarised in Table \ref{tab:model_comparison}.

\begin{table}[htbp]
\centering
\small
\setlength{\tabcolsep}{3pt}
\renewcommand{\arraystretch}{1.15}
\caption{\textbf{Comparison of models for uncertainty reducibility and explainability analysis.} Both models use default \texttt{scikit-learn} settings~\cite{pedregosa2011scikit}, with $n_{\text{estimators}}=500$ for Random Forest and standard scaling for Ridge Regression.}

\label{tab:model_comparison}
\begin{tabular}{p{3.6cm} p{3.6cm} p{3.6cm}}
\toprule
\textbf{Aspect} & \textbf{Random Forest\cite{breiman2001random}} & \textbf{Ridge Regression\cite{hoerl1970ridge}} \\
\midrule
Model type &
Non-linear ensemble &
Linear parametric \\

Inductive bias &
Low bias, high flexibility &
Strong bias, constrained complexity \\

Heterogeneity handling &
Captures local and regime-specific patterns &
Assumes global linear structure \\

Localization sensitivity &
High sensitivity to calibration partitions &
Moderate, smoother response \\

\bottomrule
\end{tabular}
\end{table}

\subsubsection{\textbf{Evaluation Setup}}

We evaluate the proposed framework in four steps.
First, we employ the local model instability proxy $\sigma^*$
(Definition~\ref{sigma}) to capture instance-wise \emph{model-dependent epistemic fragility}, reflecting sensitivity due to limited training support, model instability, or structural misspecification. This proxy is fixed prior to calibration localisation and does not depend on the localisation trajectory. 

Second, we introduce correlation-based diagnostic metric, \emph{Reducibility Instability Alignment} (RIA, Definition~\ref{era})  to assess
whether calibration-induced uncertainty reducibility behaves consistently with independent signals of local model instability at the instance level. 

\begin{definition}[Reducibility Instability Alignment (RIA)]
\label{era}
It measures the extent to which
\emph{calibration-induced uncertainty reducibility} behaves consistently with
instance-wise \emph{model instability}. It is defined via Pearson correlation \cite{pearson1895vii} as
\begin{equation}
\mathrm{RIA}_{W} = \mathrm{corr}_{\mathrm{Pearson}}(\sigma^*, W_{\mathrm{red}}),
\qquad
\mathrm{RIA}_{R} = \mathrm{corr}_{\mathrm{Pearson}}(\sigma^*, R_{\mathrm{red}}),
\end{equation}
capturing the association between local model instability and the absolute
or the relative magnitude of uncertainty inflation induced by global calibration and revealed through localisation.
\end{definition}
High \textit{RIA} values link local instability to increased reducible uncertainty under global calibration; \textit{RIA} characterises association, not epistemic uncertainty. In particular, it addresses the following diagnostic question:
\begin{quote}
\emph{``Does higher model instability lead to greater calibration-induced reducible uncertainty?''}
\end{quote}

Third, we also analyse robustness to localisation resolution using granularity-based measures, which quantify the stability of inferred diagnostic relationships across different
localisation granularities.

Finally, the last step is to compare \textit{ConformaDecompose} with the baselines introduced in Table \ref{tab:comparison}.

%To quantify how epistemic uncertainty relates to reducible uncertainty, we introduce two complementary correlation-based alignment metrics that capture both linear and monotonic dependencies.

\begin{definition}[Granularity Sensitivity (GS)]
\label{gs}
Let $K_1$ and $K_2$ denote two localisation granularities.
For a metric $M \in \{\mathrm{RIA}_W,\mathrm{RIA}_R\}$, the \emph{granularity sensitivity} is defined as
\begin{equation}
\mathrm{GS}_{M} = \left| M^{(K_2)} - M^{(K_1)} \right|.
\end{equation}

\end{definition}

Low values of \textit{GS} indicate stability of the inferred relationship between epistemic uncertainty and reducible uncertainty with respect to localisation granularity, while high values indicate sensitivity to neighbourhood resolution. It addresses the following question:

\begin{quote}
\emph{``Does the inferred alignment between epistemic uncertainty and reducible uncertainty remain stable when the localisation granularity changes?''}
\end{quote}

\subsubsection{\textbf{Implementation}}

To assess the effect of localisation granularity on uncertainty reduction, all experiments are conducted using two clustering configurations on the calibration set: $K=5$ (coarse localisation) and $K=8$ (finer localisation). 
These values are selected to balance expressive neighbourhood resolution with calibration stability, ensuring sufficient calibration support for the evaluation datasets.

For each dataset, model, and clustering configuration, \emph{ConformaDecompose} is applied to all test instances to compute local epistemic uncertainty $\sigma^*$, absolute and relative reducible uncertainty ($W_{\mathrm{red}}, R_{\mathrm{red}}$), and their alignment metric (RIA, Definition \ref{era}). Dataset-level summaries are obtained by correlating these quantities across the test set. GS (Definition \ref{gs}) measures the absolute change in alignment metrics between $K=5$ and $K=8$, assessing the stability of inferred uncertainty behaviour with respect to localisation resolution. Pearson correlation captures proportional association, whereas Spearman's rank correlation assesses monotonic dependence and robustness. Strong alignment between $\sigma^*$ and $W_{\mathrm{red}}$ indicates epistemic control over uncertainty magnitude, whereas alignment with $R_{\mathrm{red}}$ reflects the relative contribution of epistemic uncertainty.

Finally, \emph{ConformaDecompose} is evaluated against local and weighted conformal baselines (Section~4.1), enabling comparison in terms of the interpretability introduced by calibration localisation, addressing the following question:
\begin{quote}
  \emph{``Does calibration localisation enable interpretable calibration-induced conformal uncertainty reduction beyond local and weighted baselines?''}  
\end{quote}

\section{Results and Discussion}

\textbf{Local Model Stability Alignment.}
Table~\ref{tab:era_ria_merged} reports RIA (Definition~\ref{era}) across datasets, models, and localisation granularities, providing empirical evidence that calibration-induced reducible uncertainty covaries with local model instability.

\begin{table*}[htbp]
\centering
\small
\setlength{\tabcolsep}{3.5pt}
\renewcommand{\arraystretch}{1.15}
\caption{\textbf{Instance-wise verification of reducible uncertainty (calibration-induced) and sensitivity to localisation granularity.}
Values report Pearson correlations (RIA, $p<0.01$) between local model instability and reducible width ($RIA_{\mathrm{W}}$) and reducible ratio ($RIA_{\mathrm{R}}$) at $K=5,8$.
GS (Definition \ref{gs}) measures the absolute change in RIA between $K=5$ and $K=8$. Minimum-weight $\varepsilon$ is set to $1e-3$.}
\label{tab:era_ria_merged}

\begin{tabular}{llcccccccc}
\toprule
& & \multicolumn{2}{c}{\textbf{K = 5}} & \multicolumn{2}{c}{\textbf{K = 8}}
& \multicolumn{2}{c}{\textbf{GS ($|K{=}8 - K{=}5|$)}} \\
\cmidrule(lr){3-4} \cmidrule(lr){5-6} \cmidrule(lr){7-8}

\textbf{Dataset} & \textbf{Model}
& $RIA_{\mathrm{W}}$ & $RIA_{\mathrm{R}}$
& $RIA_{\mathrm{W}}$ & $RIA_{\mathrm{R}}$
& $\mathrm{GS}_{RIA_{\mathrm{W}}}$ & $\mathrm{GS}_{RIA_{\mathrm{R}}}$ \\
\midrule

Power P.
& RF & 0.99 & 0.90 & 0.82 & 0.74 & 0.03 & 0.02 \\
& RR & 0.73 & 0.49 & 0.58 & 0.17 & 0.15 & 0.32 \\

Parkinson
& RF & 0.99 & -0.09 & 0.95 & -0.26 & 0.05 & 0.17 \\
& RR & 0.46 & 0.36 & 0.87 & 0.23 & 0.41 & 0.13 \\

Airfoil
& RF & 0.97 & -0.04 & 0.99 & 0.10 & 0.03 & 0.14 \\
& RR & 0.99 & 0.09 & 0.88 & 0.10 & 0.12 & 0.01 \\

Abalone
& RF & 0.98 & 0.09 & 0.97 & 0.56 & 0.03 & 0.47 \\
& RR & 0.99 & 0.63 & 0.91 & -0.60 & 0.08 & 1.23 \\

\bottomrule
\end{tabular}
\end{table*}

Across datasets and models, reducible uncertainty aligns strongly with local model instability in \emph{absolute magnitude}, as indicated by consistently high correlations with reducible width ($RIA_{\mathrm{W}}$, typically $>0.9$). In contrast, alignment with the reducible ratio ($RIA_{\mathrm{R}}$) is more variable, ranging from negative to strongly positive values. This divergence verifies that calibration-induced epistemic uncertainty scales reliably with model instability in magnitude, while its \emph{relative contribution} depends on irreducible uncertainty and localisation granularity. Strong width alignment thus confirms epistemic consistency, whereas variability in ratio alignment reflects attenuation effects due to noise and resolution.

Accordingly, high-width alignment with weak or negative ratio alignment indicates calibration-induced epistemic uncertainty diluted by irreducible components, while strong alignment in both measures reflects substantial structural reducibility. Intermediate cases arise when ratio alignment varies across granularities, indicating resolution-sensitive reducibility rather than inconsistency. Consistent with these verification patterns, dataset-specific behaviours align with known ground-truth characteristics and are summarised below.

%Across all settings, width-based metrics remain stable and consistently high, supporting robust uncertainty contraction under localization, whereas ratio-based metrics exhibit greater variability. This contrast highlights the distinction between \emph{absolute} and \emph{relative} reducibility and reveals model- and dataset-dependent attenuation effects. 

\begin{itemize}
    \item \textbf{Power Plant}: Characterized by strong physical structure and moderate sensor noise~\cite{tufekci2014prediction}. Reducible uncertainty shows strong epistemic consistency in magnitude, with relative reducibility improving under finer localisation, indicating recoverable structure once coarse noise is controlled.
    
    \item \textbf{Parkinson}: Dominated by biological variability and measurement noise~\cite{little2009parkinsons}. Reducible uncertainty scales in magnitude with model instability, while relative reducibility remains limited across granularities, consistent with dominant irreducible components.
    
    \item \textbf{Airfoil}: A high-noise aerodynamic dataset~\cite{brooks1989airfoil}. Strong magnitude alignment but weak relative alignment indicate that epistemic effects are largely masked by experimental noise, even at finer localisation.
    
    \item \textbf{Abalone}: A heterogeneous benchmark with known model mismatch~\cite{abalone_1}. Reducible uncertainty exhibits strong epistemic consistency, with relative reducibility increasing under finer localisation and signs of resolution sensitivity at higher granularities.
\end{itemize}

\noindent
\textbf{Sensitivity to Localisation Granularity.}
Table~\ref{tab:era_ria_merged} also reports GS (Definition~\ref{gs}) of RIA when increasing localization from $K=5$ to $K=8$. Width-based metrics exhibit consistently low sensitivity, indicating stability with respect to localisation resolution, whereas higher sensitivity in ratio-based metrics (e.g., \emph{Power Plant}–RF, \emph{Abalone}–RR) highlights boundary cases where reducible composition depends on neighbourhood resolution. Overall, uncertainty reducibility remains robust to localisation granularity: width-based GS values are typically small ($<0.05$), while larger but localised variations in ratio-based GS (occasionally exceeding $0.5$) reflect compositional sensitivity rather than changes in the underlying uncertainty regime.

\noindent
\textbf{Comparison with Local and Weighted Baselines.}  Table~\ref{tab:width_rr_cov} reports a comparison between \textit{ConformaDecompose} and local and weighted conformal baselines in terms of interval width, empirical coverage, and reducible uncertainty.
The results show that \textit{ConformaDecompose} enables an interpretable decomposition of conformal uncertainty by explicitly quantifying the portion of interval width that is structurally reducible through calibration localisation.

\begin{table}[htbp]
\centering
\caption{\textbf{Comparing \textit{ConformaDecompose} to local and weighted conformal baselines.}
$W$ denotes the average prediction interval width before normalisation (lower is better).
RR denotes the dataset-specific reducible ratio and is only defined for \textit{ConformaDecompose}.
Cov denotes the empirical coverage (in \%). $\alpha$ denotes the miscoverage level and is set to 0.10. $K$ is set to 5 for comparability in all experiments. }
\label{tab:width_rr_cov}
\setlength{\tabcolsep}{3pt}
\begin{tabular}{l|ccc|ccc|ccc|ccc}
\hline
 & \multicolumn{12}{c}{\textbf{Dataset}} \\
\textbf{Method }
& \multicolumn{3}{c}{Airfoil}
& \multicolumn{3}{c}{Abalone}
& \multicolumn{3}{c}{Parkinson}
& \multicolumn{3}{c}{PowerP.} \\
\cline{2-13}
 & W & RR & Cov
 & W & RR & Cov
 & W & RR & Cov
 & W & RR & Cov \\
\hline
Normalized   & 7.5 & -- & 89\% & 6.8 & -- & 89\% & 1.5 & -- & 89\% & 10.9 & -- & 89\% \\
CQR          & 7.5 & -- & 90\% & 7.0 & -- & 91\% & 1.2 & -- & 89\% & 10.9 & -- & 90\% \\
UACQR-S        & 7.4 & -- & 89\% & 7.0 & -- & 91\% & 1.2 & -- & 89\% & 11.0 & -- & 90\% \\
%EPISC.       & 7.7 & -- & 90\% & 7.0 & -- & 91\% & 1.2 & -- & 89\% & 10.9 & -- & 90\% \\
Mondrian     & 7.5 & -- & 87\% & 6.8 & -- & 89\% & 5.5 & -- & 89\% & 10.0 & -- & 88\% \\
\hline
\textbf{CD (glob)}  
& \textbf{7.8} & \textbf{--} & \textbf{89\%}
& \textbf{6.7} & \textbf{--} & \textbf{91\%}
& \textbf{5.6} & \textbf{--} & \textbf{89\%}
& \textbf{10.0} & \textbf{--} & \textbf{89\%} \\
\textbf{CD (min)}   
& \textbf{7.0} & \textbf{0.1\%} & \textbf{87\%}
& \textbf{5.7} & \textbf{1.0\%} & \textbf{85\%}
& \textbf{4.9} & \textbf{0.1\%} & \textbf{87\%}
& \textbf{9.3} & \textbf{0.7\%} & \textbf{87\%} \\
\hline
\end{tabular}
\end{table}

Across datasets, reductions of up to 1.0\% are achieved while maintaining near-nominal coverage, showing that part of conformal uncertainty is systematically interpretable and reducible beyond existing local or weighted methods. \textit{ConformaDecompose} preserves marginal validity while exposing instance-level calibration-induced reducible uncertainty. In contrast, apparent gains of some baselines (e.g., Parkinson) largely arise from aggressive interval shrinking that removes aleatoric uncertainty rather than addressing epistemic reducibility.

\section{Conclusions, Limitations, and Future Work}

This work introduces \emph{ConformaDecompose}, a framework for instance-wise empirical decomposition of predictive uncertainty into calibration-induced epistemic (reducible) and irreducible components, built on the robust uncertainty quantification principles of conformal prediction. \emph{ConformaDecompose} delivers a four-part explanation that makes model uncertainty interpretable to end users. It illustrates (i) the uncertainty localisation trajectory, increasing transparency of the calibration process; (ii) the difference between global and localised conformal interval widths; (iii) the evolution of calibration support as specific clusters are progressively downweighted; and (iv) the cluster-wise contribution to uncertainty reduction, identifying which calibration regions drive interval contraction. Unlike standard feature importance or attribution plots, \emph{ConformaDecompose} adopts a user-centric perspective by explicitly communicating model uncertainty. By making uncertainty visible and interpretable, it enables end users to understand in which settings predictions can be trusted, up to which point calibration-induced uncertainty remains reducible, and beyond which point the data should be monitored or controlled due to irreducible, data-related uncertainty.

ConformaDecompose reveals calibration-induced epistemic uncertainty and its reducibility but provides no conformal guarantees and does not capture uncertainty from model misspecification or limited capacity. It serves as a diagnostic tool for locating uncertainty to guide model refinement; residual uncertainty reflects irreducible noise and model limits. Reducibility patterns depend on model-derived uncertainty proxies and clustering-based localisation, and all alignments are associative rather than causal.

Future work will explore adaptive and data-driven localisation mechanisms, extensions to sequential and non-stationary settings, and integration with active learning and uncertainty-aware decision-making. Importantly, the proposed approach builds on conformal prediction, which is inherently model-agnostic and can be applied on top of any predictive model. The specific use case and learning task are determined by the choice of the nonconformity score, enabling straightforward extensions beyond the forecasting setting considered in this work to problems such as classification, regression, and other predictive tasks. More broadly, the concept of uncertainty reducibility offers a promising direction for distinguishing intrinsic noise from resolvable model uncertainty, supporting the development of reliable and safety-critical predictive systems.

\section{Acknowledgments}
Fatima Rabia is a PhD student at the University of Bologna (DISI), funded by PNRR (Piano Nazionale di Ripresa e Resilienza) and Automobili Lamborghini S.p.A., Italy. She thanks Dr.~Juan Carlos Andresen and Abhishek Srinivasan from TRATON AB for valuable discussions on uncertainty quantification and conformal prediction, and Şakir Furkan Yöndem, and Gizem Karagöz for discussions on evaluation and review support. Helena Löfström acknowledges support from the Kamprad Family Foundation (grant 20250309). Tuwe Löfström acknowledges support from the Swedish Knowledge Foundation, industrial partners, and Jönköping University, Sweden (PREMACOP, grant 20220187).

\section{Disclosure of interests}
We declare no competing interests for this study.

\bibliographystyle{unsrt}

\bibliography{pmlr-sample}

\end{document}